\documentclass[conference,letterpaper]{IEEEtran}

\usepackage{amsfonts}
\usepackage{graphicx,times,amsmath} 
\usepackage{latexsym}
\usepackage{amssymb}
\usepackage{color}

\IEEEoverridecommandlockouts

\definecolor {grey}{rgb}{0.2,0.2,0.2}
\definecolor {dkgr}{rgb}{0,0.3,0}

\begin{document}

\title{\ \\ \LARGE\bf The Riddle of Togelby}
\author{Daniel Ashlock, Senior Member IEEE and Christoph Salge, Member
  IEEE \thanks{Daniel Ashlock is at the University of Guelph,
    dashlock@uoguelph.ca} \thanks{Christoph Salge is at the University
    of Hertfordshire,christophsalge@gmail.com} \thanks{The authors
    thank the Canadian Natural Sciences and Engineering Research
    Council of Canada (NSERC) and the European Commission (Grant INTERCOGAM) for supporting this work.}}

\date{}

\maketitle

\begin{abstract}
At the 2017 Artificial and Computational Intelligence in Games meeting at Dagstuhl, Julian Togelius asked how to
make spaces where every way of filling in the details yielded a good
game.  This study examines the possibility of enriching search spaces
so that they contain very high rates of interesting
objects, specifically game elements.  While we do not answer the full
challenge of finding good games throughout the space, this study
highlights a number of potential avenues.  These include naturally
rich spaces, a simple technique for modifying a representation to
search only rich parts of a larger search space, and representations
that are highly expressive and so exhibit highly restricted and
consequently enriched search spaces. 
\end{abstract}

\section{Introduction}

Let us first put the main question into context. Procedural content generation (PCG) \cite{short2017procedural,katecompton2016,shaker2016procedural} is about finding algorithmic methods to generate content. We are looking specifically at PCG in games, where it is used to create a variety of things, such as levels, characters, tress, text, game rules, and even, in a somewhat aspirational fashion, whole games. There are different approaches - such as composing algorithms that produce only acceptable content. As an example, imagine you want to generate some text for a non-player character. You could just define a sentence snipped, such as \textit{``Bring me XXX''} and then replace XXX from a list of things, such as \textit{(a shrubbery, the holy grail, 10 Defias Brotherhood headbands)}. These approaches usually rely on some form of grammar or regular expression, and if defined well, only produce acceptable content. See for example \cite{compton2015tracery} or the whole range of more complicated but still compositional approaches to computation narrative \cite{kybartas2017survey}. The downside of those approaches is their often limited exploration of possible options. An alternative are generate-and-test methods, also referred to as search based PCG \cite{TogeliusSBPCG}. For those you need a function that can evaluate the produce content, and tell if its acceptable, or even better, how good it is. Given such a function it is then possible to produce a wider range of more experimental content, and keep those that meet the requirements, or just keep the best. Going back to our previous example, you might make sentences for your NPC by stringing random letter from the English alphabet together. You can then test if they form a proper sentence with some grammar rules, and maybe you even have a function that tells you how good a certain sentence is. One problem with this approach is that you have no guarantee how easy it is to find a feasible solution. What we do know is that the representation of your generator matters \cite{ashlock2016representations}. Consider instead of stringing together random letters you string together random words from the English dictionary. While this limits the range of sentences you generator can produce, it also massively enhances the change to find an acceptable one. We know from previous research that the right representation can influence what solutions you are likely to find, and how quick you are likely to find them \cite{Ashlock:2005:ICR:1068009.1068018}. It is in this context that The Riddle of Togelby was posed: ``Can you define a representation where every way of filling it in yields a good game?''

While this is trivially true - one could design a search space so small that it just contains one (good) game - we do not currently provide a satisfactory answer to the spirit of this question. We mainly look at systems that provide parts of games, rather than full games, and we mostly look at approaches that increase the chances of finding good solutions, rather than guaranteeing it. What we do provide is a list of example that contain different approaches to finding or designing such search spaces, and we also discuss the properties or tricks used to make them. 

One common theme here is the trade off between expressivity of the space, i.e. how different are the things the generator produces, vs. your chance of finding a good solution. We will not look at this quantitatively here, as one could with an expressive range analysis \cite{smith2010analyzing}. We rather wanted to give an overview of different ways one could push to the Pareto front of this trade-off - looking for solutions that offer the best of both. We also want to discuss how certain representation achieve this, and take a look on how well those deal with adaptation to existing content and control from a human user. While we are interested in presenting the representation within this document, the vision for this paper is to figure out strategies on how to address the riddle of Togelby in its entirety.

\subsection{Overview}

The rest of this paper will present different examples for generative systems, and discusses their properties. 

We first present a naturally rich representation that is highly controllable base on Voronoi
tilings \cite{Aurenhammer91}. Then, we look at a search space that can be restricted
to make it naturally rich for generating apoptotic cellular
automata \cite{Ashlock13sp1}.  These examples are not generalizable
except by a process of discovery, they contain no overarching
principle that makes the spaces rich, but it is still an interesting
example of a rich search space.

Beyond having a naturally rich space a technique called {\em single
  parent crossover} is presented that can be used to focus search in a rich part of the
space.  The effect of single parent crossover is to focus search in
the part of gene-space where examples lie.  The technique acts by
permitting immortal population members, which are the examples,
participate in a one-sided form of crossover \cite{Ashlock12sp,
  Ashlock05sp}.

Finally a new technique called {\em convex representation} is presented, in which we can pick out a small subset of a search space and use an alternate representation to search only that part of the space.  A representation is convex if the weighted average of instances of a representation are, themselves, instances of the representation.  This permits search to be restricted, in a transparent fashion, to the convex hull of a collection of examples.

\section{Related Work
}


Automated level generation in video games can arguably be traced back
to a number of related games from the 1980s (Rogue, Hack, and
NetHack), collectively called {\em Roguelike games}.  The task is
currently of interest to the research community.  In
\cite{Sorenson10a} levels for 2D sidescroller and top-down 2D
adventure games are automatically generated using a two population
feasible/infeasible evolutionary algorithm. Answer set programming is
another approach to dungeon generation~\cite{smith2014logical}. In
\cite{Julian10a} multiobjective optimization is applied to the task of
search-based procedural content generation for real time strategy
maps.

\begin{figure}[htb]
\centerline{
\includegraphics[width=0.48\textwidth]{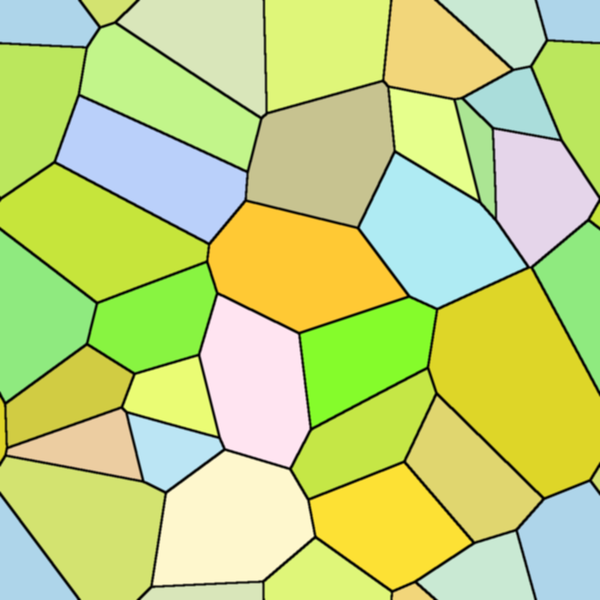}
}
\vspace{0.04in}
\centerline{
\includegraphics[width=0.48\textwidth]{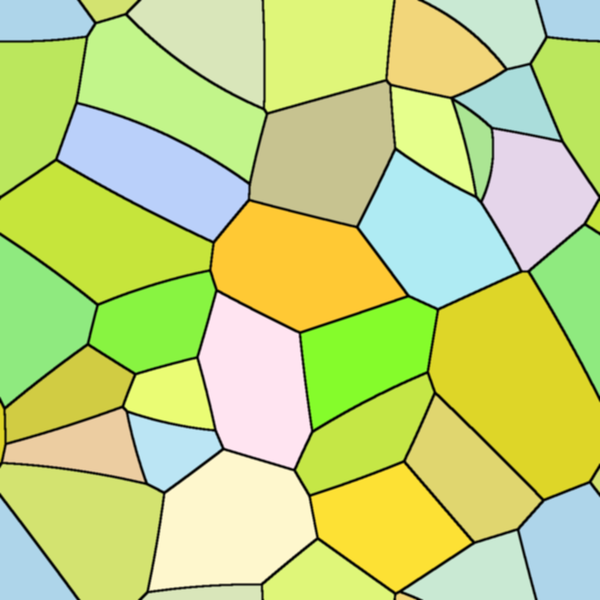}
}
\caption{The upper panel shows a standard Voronoi diagram, the lower
  one a weighted Voronoi diagram with curved tile sides.}
\label{exVT}
\end{figure}

\section{Naturally rich spaces}
First, we will look at an example from a naturally rich space - by that we mean representations where nearly all examples from the space a feasible solutions. 
\label{NRS}
\subsection{Voronoi Tiling}
\begin{figure*}[tb]
\centerline{
\includegraphics[width=0.3\textwidth]{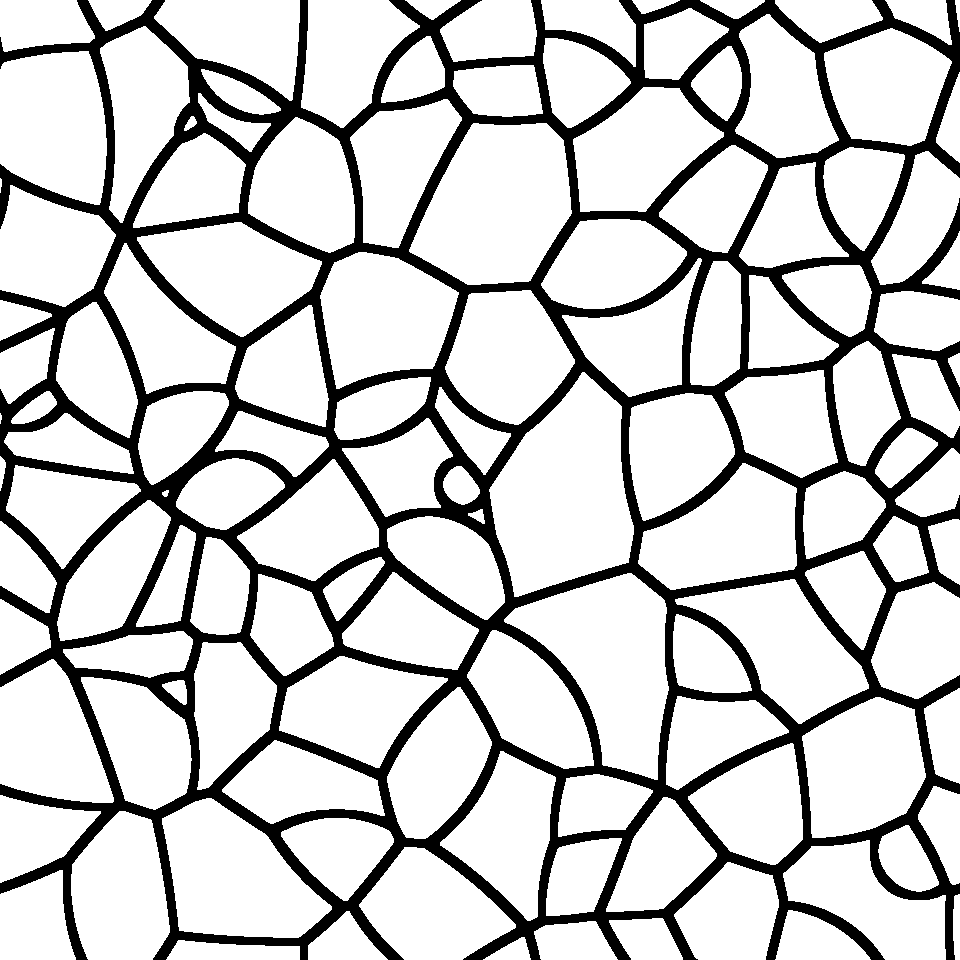}
\hspace{0.02\textwidth}  
\includegraphics[width=0.3\textwidth]{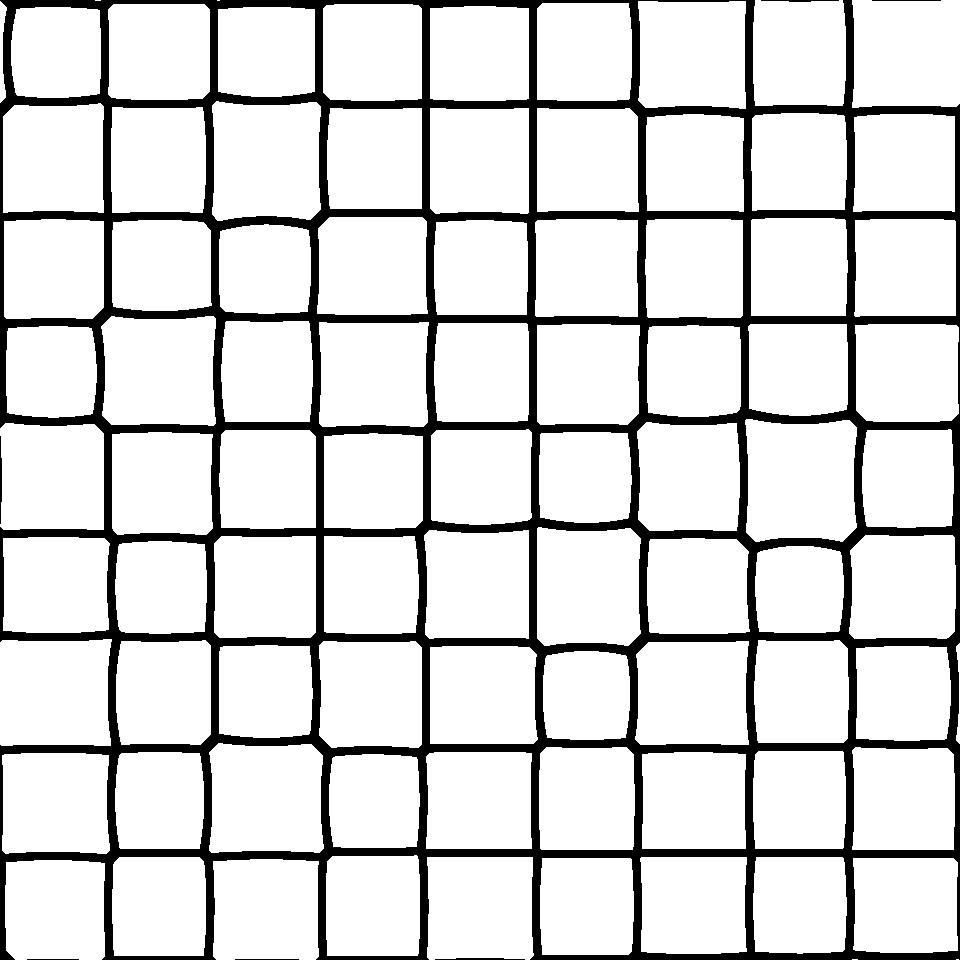}
\hspace{0.02\textwidth}  
\includegraphics[width=0.3\textwidth]{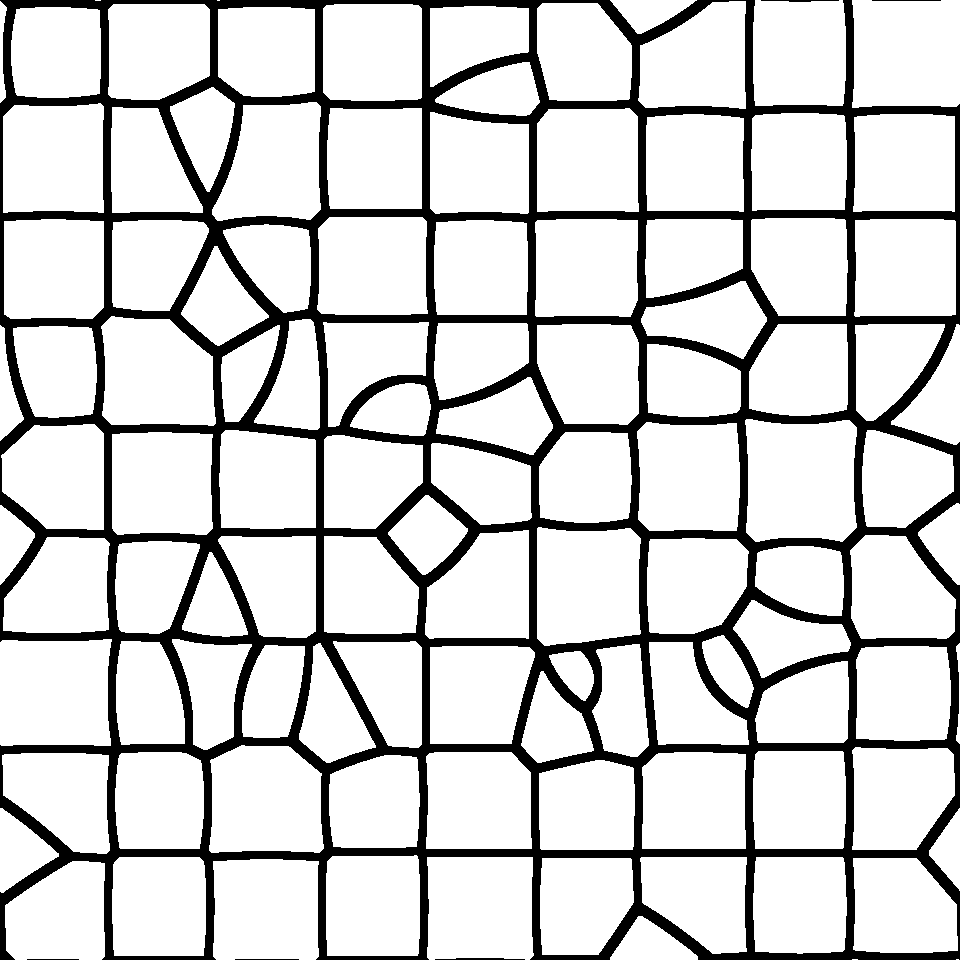}
}
\caption{Shown are examples of weighted Voronoi street maps using random points (left), a square grid of points (middle) and a square grid of points with 19 added random points (right).}
\label{Streets}
\end{figure*}

A problem that arises from time to time is laying out streets for a procedurally generated village or town. We begin with the generation of a system of connected paths like those shown in Figure \ref{Streets}.  A Voronoi diagram \cite{Aurenhammer91} is created by choosing a collection of points, called tile centers, in
the plane and then creating tiles by declaring the tile associated with one of the points to consist of the parts of the plane closer to that tile.  Two sorts of Voronoi diagrams are shown in Figure \ref{exVT}.  The upper one is a standard tiling, the lower one is modified by weighting the individual points and defining tales based on distance times weight; this makes the boundaries quadratic curves instead of strait lines, giving the tiles curved sides.

Figure \ref{Streets} show three examples of layouts for the roads in a
segment of a procedurally generated village.  These are the
boundaries between Voronoi tiles.  If the weighting factors are not
too different from one another, then the resulting street map is
connected.  The three examples given in Figure \ref{Streets} show that
these layouts are easily controllable.  The left-most example uses 100
random points.  The middle uses a regular $9\times 9$ grid of points
while the right-most adds 19 random points to the 81 regular points
from the middle example.

This space of layouts for streets always yields a connected set of
paths, the shape can be controlled by the choice of tile centers, and
could be procedurally refined if specific criteria are required.  In
any case this {\em Voronoi path network} generation scheme is the
first example of a naturally rich space.  If complex requirements were
needed for a given game, then the space of Voronoi paths could be
procedurally searched for refined designs, but it starts eminently rich
in good structures. 

The question is now, why is the Voronoi tiling such a good representation? One answer to this question lies in the specific genotype-phenotype mapping of the Voronoi representation. The genome, i.e. the parameters of the actual search space, are just positions of points, yet they get translated into a phenotype, the actual tiling seen in the figures, in a relatively complex process, as described above. First, this process is very similar to several processes in nature, leading to a certain familiarity. Voronoi tiling is a particular good fit if the artefact generated in question, such as a street map, could in fact be generated by some self-organizing process based on distance based tiling in the real world.  

The same process also allows for a certain degree of control over the output. The distance function used can for example be weighted to create curved lines. One could even use it to adapt the output to existing content. Imagine, for example, that they points where in a real existing map, and the tiling would be some kind of separation in nation states. The distance function could then be based on infrastructure, i.e. existing roads, etc., which would then lead to an even richer tiling that adapts to some underlying property. 

So, the Voronoi approach is a rich representation, that allows for decent control and adaptive, but is best used to generate things that arise from a similar process as it complex genotype-phenotype mapping. One approach to duplicate its success would be to look at representations that use a process similar to the process generating the artefact, i.e. rather than generating the final artefact one might try to find a mathematical representation of how this thing came about. 

\subsection{Apoptotic Cellular Automata}

Another example of a naturally enriched space is that of the apoptotic
cellular automata described in \cite{Ashlock13sp1}.  CAs have been
applied to the visual arts, used to produce artistic images\cite{Dan1,
  art}, and their use has been extended to the fields of architecture
and urban design\cite{arch1, arch2}. An interesting application has
been the use of CA in simulating the emergence of the complex
architectural features found in ancient Indonesian structures, such as
the Borobudur Temple\cite{indo}. Ashlock and Tsang\cite{Dan1} produced
evolved art using 1-dimensional CA rules. CA rules were evolved using
a string representation. The CA either underwent slow persistent
growth, or planned senescence. The resulting fitness landscapes were
rugged with many local optima.

\begin{figure*}[tb]
\centerline{\includegraphics[width=0.85\textwidth]{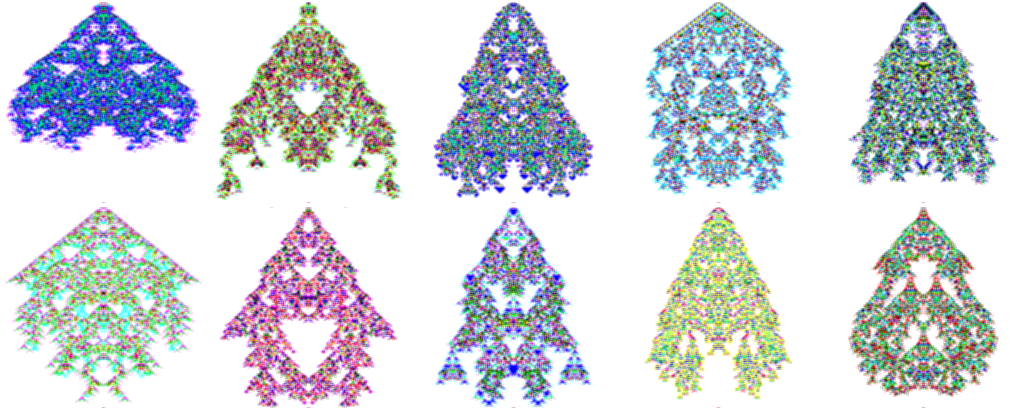}}
\caption{Ten examples of renderings of evolved apoptotic cellular automata.}  
\label{CAex}
\end{figure*}

An {\em apoptotic cellular automaton} is one that, for a given initial
condition, fills space and then ``dies'', i.e. returns all cells to a
quiescent state.  The fitness of an automata is the space it fills,
save that it is awarded zero fitness if it grows too long.  Examples
of these automata appear in Figure \ref{CAex}.  In
\cite{Ashlock13sp1}, random transects of the fitness landscape for
these automata were investigated.  This yielded a startling discovery.
A random transect is taken by starting with a high-fitness automata
rule, represented by an array of integers, and then changing the
entries, in a random order, to those of another rule.  The
intermediate rules so generated form the points in the transect.  The
fitness of automata in the transects were at least two hundred times
as large as the fitness of randomly sampled automata.

Again here the process that maps the actual genome, the relatively compact starting condition, to the outcome, is a complex process that is also used to model a range of natural phenomena. Again, the process can be manipulated to adapt the outcome to underlying properties of the environment, and we can even imagine setting some those cells to fixed values beforehand to simulate the integration of existing content. 

This is not an example replicable outside of the apoptotic cellular
automaton space, though it does enable an example of one of the
enriched representations in the next section.  The full space of
apoptotic cellular automata rules is extremely poor in good optima,
but, for reasons we do not understand, there is a huge cluster of
them in one part of the fitness landscape.  The rules in this paper
were 36 dimensional, with one constraint -- that the zeroth entry
must be zero -- to enable death.  This means that there are $2^{35}$
random transects between any two high fitness rules, of which hundreds
were located.  This gives us a very rich search space for these images
that could be used as aliens in a space-invader like game or enemy
ships in a game like Galaga.  These images can be generated from a 36
byte rule, permitting thousands to be pre-generated and stored in even
a small application.

An example of a highly enriched space that occurs outside of games is
that of {\em side effect machines} \cite{McEachern14a}.  A side effect
machine is a finite state machines whose transitions are driven by a
biological sequence, like DNA.  The machine counts how many times it
enters each state and the counts form features for classification.
Side effect machines are evolvable transducers for changing variable
length biological sequences into numerical features.  In the course of
the research of side effect machines it was found that evolving these
machines with a fitness function based on classification accuracy
started at high values with evolution improving the fitness by 10-20\%
over that available in the initial random population.

\section{Enriched Representations}
In this section we talk about enriched representation, those that are not naturally rich in examples, but can be modified or limited in a way that produces a rich subspace.
\label{ER}

\begin{figure*}[tb]
\centerline{\includegraphics[width=0.9\textwidth]{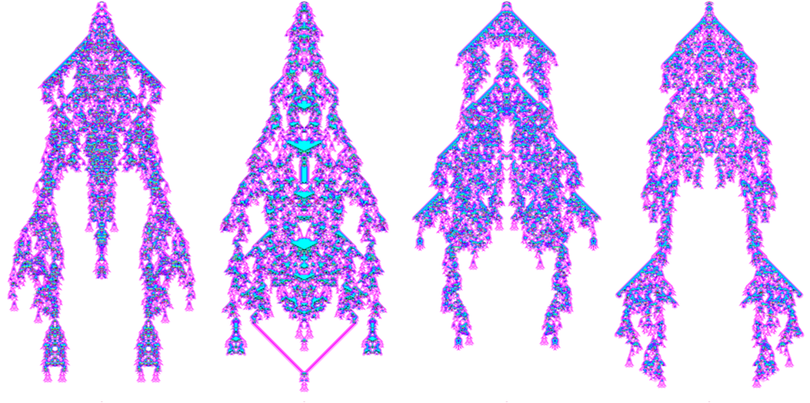}}
\caption{Examples of four renderings of apoptotic cellular automata
  derived from the same ancestor.}
\label{SP}
\end{figure*}

\subsection{Apoptotic Cellular Automata}

The apoptotic cellular automata use a {\em drawing arena}.  If their
time histories remain alive when they reach the bottom of the arena,
then they did not die in time and are awarded zero fitness.  The
automata in Figure \ref{CAex} used a $201\times 201$ arena.  A problem
with this is that these are relatively small pictures.  In
\cite{Ashlock12sp}, single parent crossover was used to generate much
larger pictures that had a similar appearance to that of a smaller
apoptotic cellular automata rendering.  The evolutionary algorithm was
modified to permit crossover with an example gene called an {\em
  ancestor} and the drawing arena was set to a much large size.

The use of single parent crossover re-injects the content of the
ancestor into the population over and over.  This makes genes similar
to the ancestor very likely and, as a side effect, restricts the
search space to points near the ancestor.  Because of the dense
packing of high fitness genes in the apoptotic cellular automata
search space, the algorithm rapidly locates high fitness genes, giving
us the desired enrichment of the search space.  Usually, though not
always, these genes are also similar in appearance to the ancestor.
This permits broad search on small examples -- which run far faster --
and then selection of a desired appearance to seed larger images through
single parent crossover.  An example of four such renderings of
apoptotic cellular automata are shown in Figure \ref{SP}.

While we have a ready example available with apoptotic cellular
automata, single parent crossover may be used to promote information
re-use and focus evolutionary search near an ancestor or ancestors
in any representation that uses crossover.  It has already been found
that single parent techniques substantially enhance the evolution of
virtual robots, using multiple ancestors, though this data have not
yet been published.

\subsection{The do what's possible representation}

A general technique for generating highly enriched search spaces is
the {\em do what's possible} representation~\cite{AshlockDWP16}.  In a
paper accepted to the CoG 2019 it was found that this representation
can lay out rooms using a tiny data structure.  The original
publication on the DWP representation~\cite{AshlockDWP16} used the
representation on a variety of problems.  It was capable of solving
self avoiding walk problems~\cite{Ashlock06a} of unprecedented size.
Where $12\times 12$ had been a practical limit with even adaptive
representations~\cite{Ashlock16agr}, the DWP representation solved
$40\times 40$ cases of the problem.  Applied to the problem of
evolving a Gray code~\cite{Wilson01} the DWP representation enabled
the evolution of an 8-bit Gray code with 256 members.  The DWP problem
was also used to evolve entropically rich binary strings and a type of
classification character for DNA in the initial study.  A later study
refined the systems power for DNA classification~\cite{AshlockDWP17}.

\begin{figure}[htb]
\centerline{\includegraphics[width=0.32\textwidth]{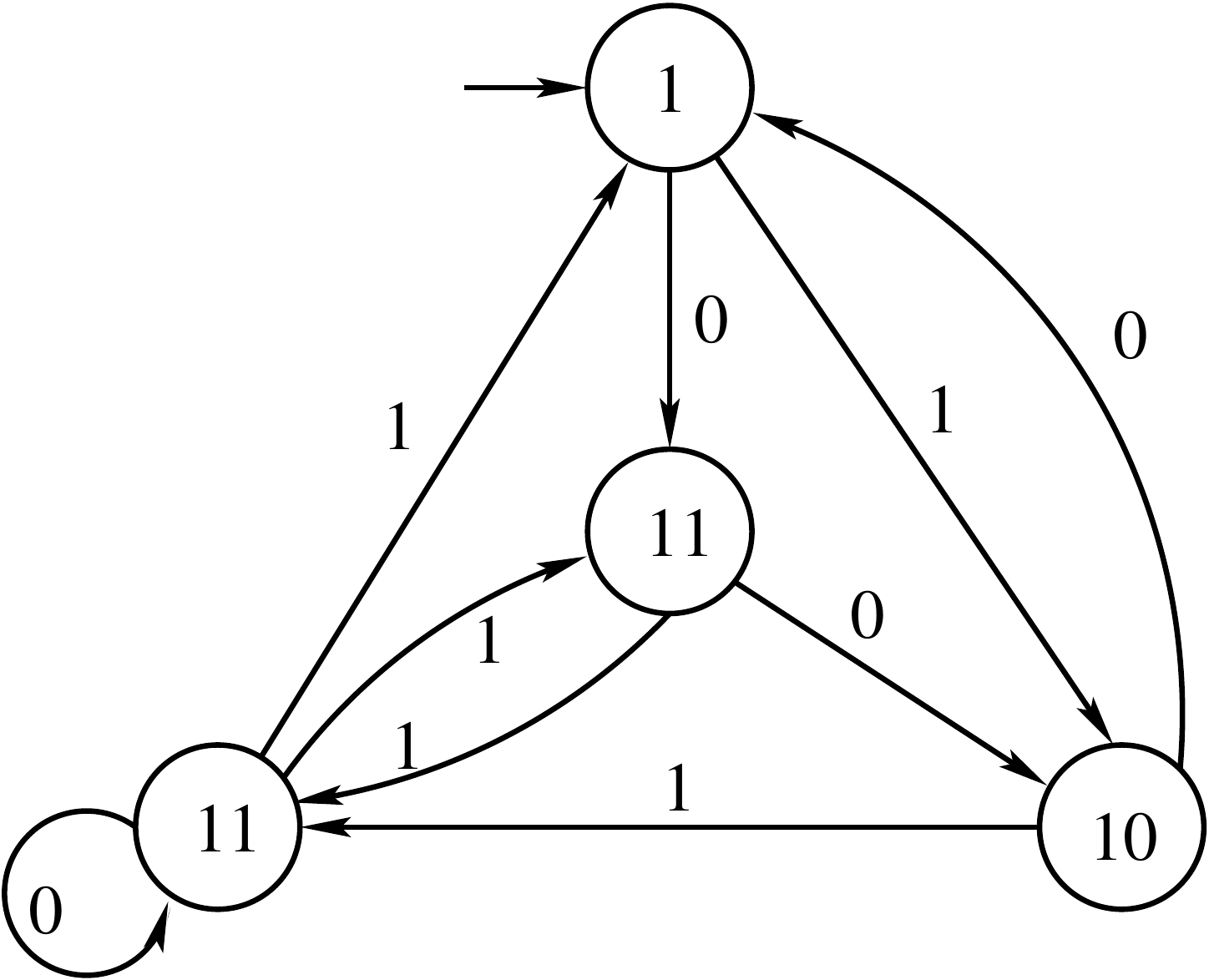}}
\centerline{\small \bf 1101111110111111...}  
\caption{An example of a self driving automata over the binary
  alphabet and the first few characters that it emits.  The sourceless
  arrow denotes the initial state. The strings emitted are shown on
  the states.  The automata's output is used as its input.}
\label{SDA1}
\end{figure}

In \cite{Ashlock19mapDWP} the representation was used to lay out large
numbers of rooms in prototype dungeon maps.  The key to this is the
use of the self-driving automata shown in Figure \ref{SDA1}.  This
data structure, driving its own transitions with its output, generates
a stream of bits that, while deterministic, can have quite a complex
pattern \cite{AshlockDWP16}.  Reading these bits in groups to generate
integers of desired size permits the self driving automata to execute
a serial decision process about where to place rooms.  The time
needed to locate good examples is short, but does not reach quite the
level of enrichment desired.

\begin{figure*}[tb]
\begin{center}
\begin{tabular}{cc}
\includegraphics[width=0.45\textwidth]{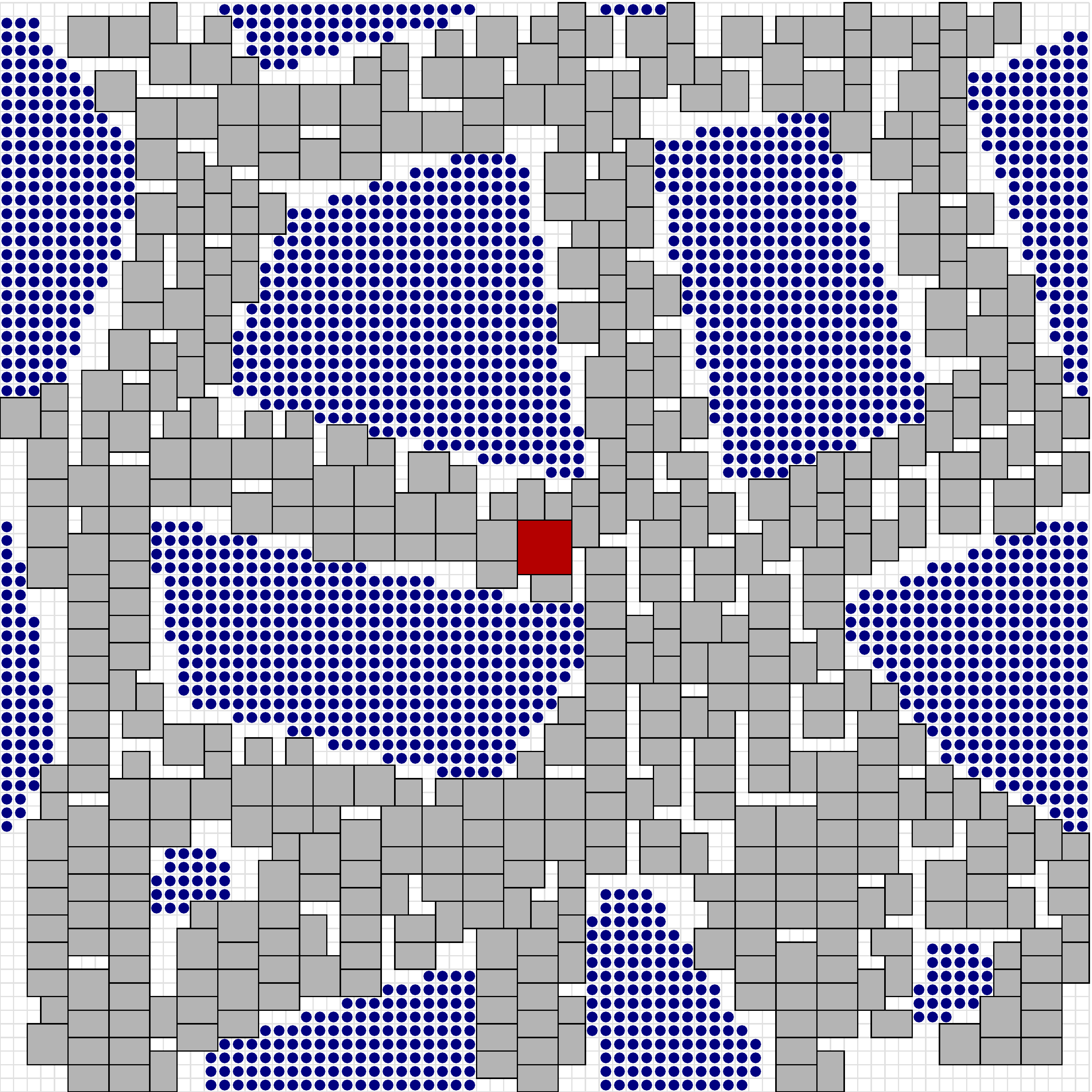}&
\includegraphics[width=0.45\textwidth]{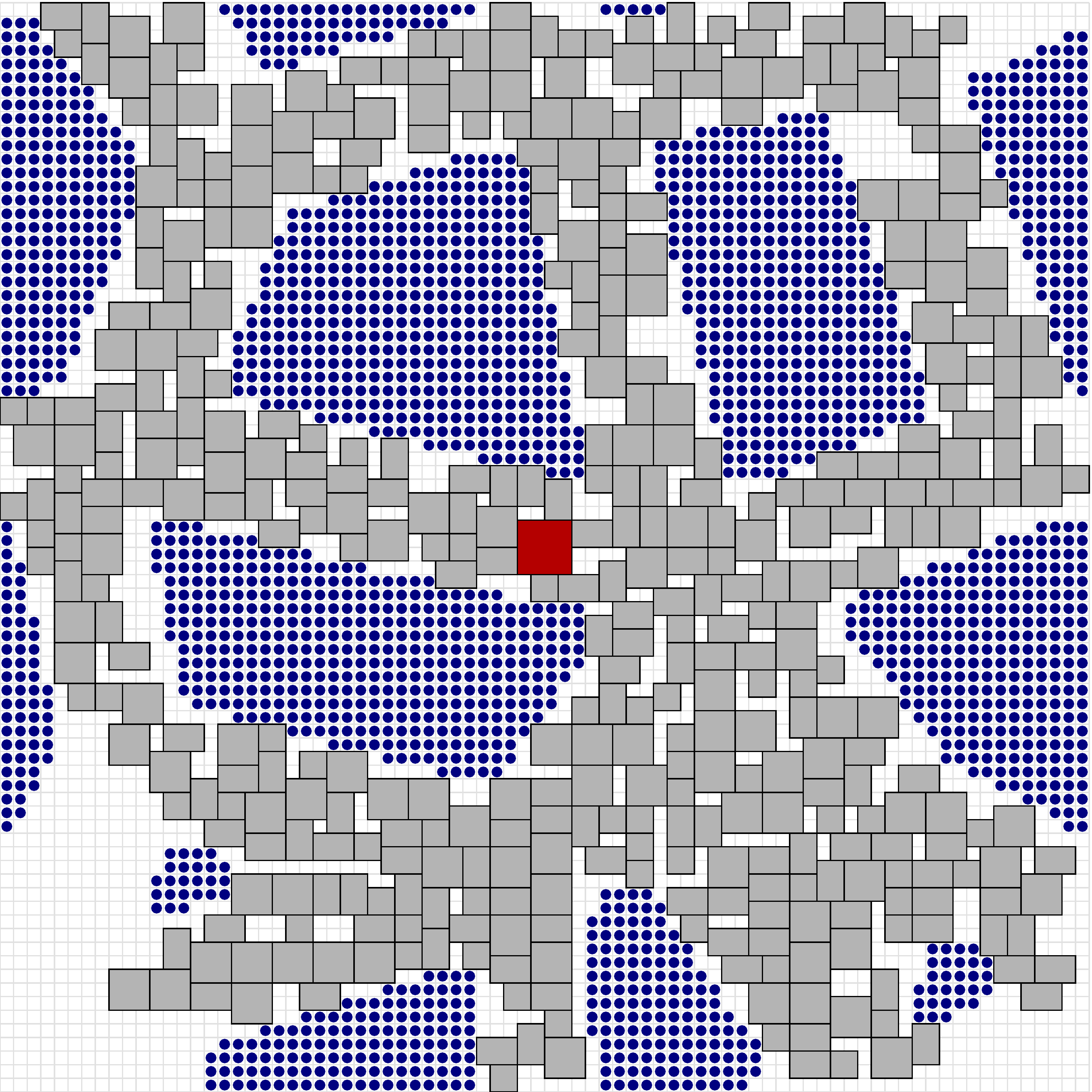}\\
\\
\includegraphics[width=0.45\textwidth]{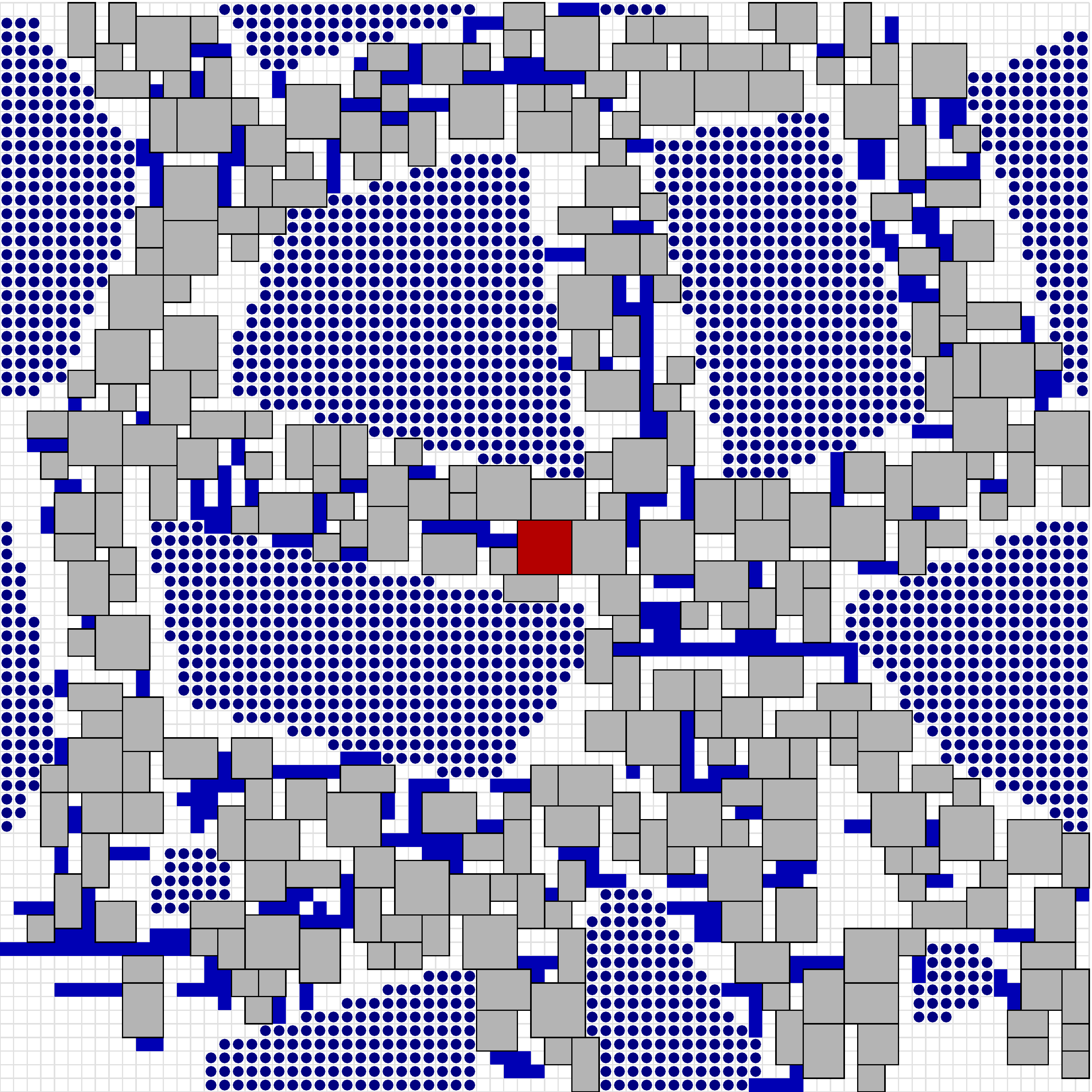}&
\includegraphics[width=0.45\textwidth]{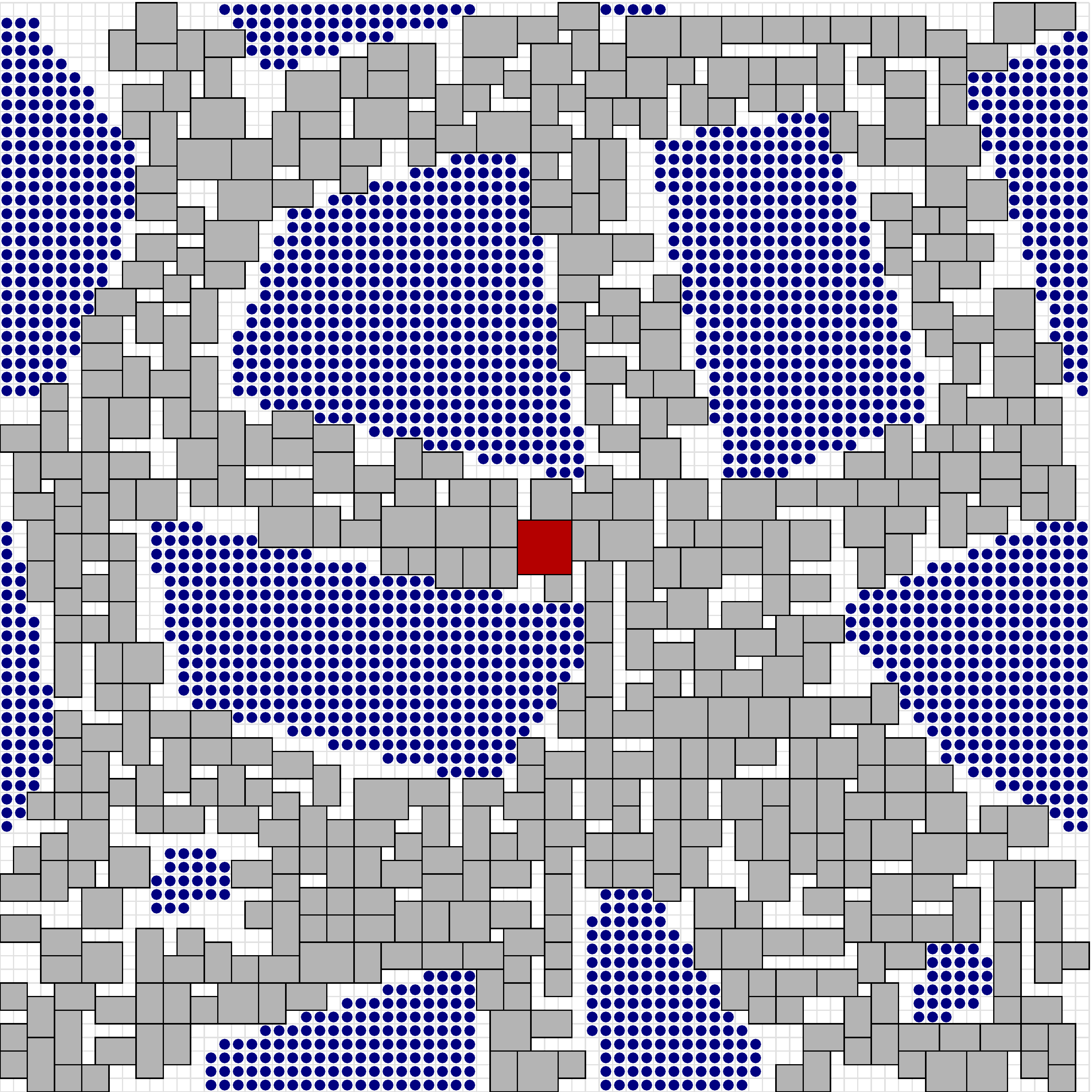}\\
\end{tabular}
\end{center}
\caption{Four different room layouts made with do-what's possible room
  generators.  Grey rectangles are rooms, blue lines are corridors.
  These room assignments were placed in a set of passages generated as
  the borders of a Voronoi diagram.  The red room is the first room
  laid down.}
\label{Dung}
\end{figure*}

The room generator, however, places rooms when it can, it ``does
what's possible'' which means a high fitness room generator is
reusable in different situations.  The self driving automata generates
a stream of proposed rooms which are laid down when they fit and
rejected when they do not.  The decision process used is the one
described in \cite{Ashlock19mapDWP}.  We used a Voronoi path network,
from Section \ref{NRS} and applied the room-layout technology.  Four
examples of layouts by different room layout engines with different
evolved self-driving automata are shown in Figure \ref{Dung}.

One modification was made in the representation for this study.  The
room selection process chooses a room by taking a large integer
supplied by the self driving automata modulo the number of existing
rooms.  This study disallows a room that has failed to have another
room placed next to it eight or more times.  This directs growth to
the active periphery of the the current layout.

In this example the enrichment is again due to manipulating the process that translates the genotype to the phenotype, but this time the feasibility filter specifically filters out actions that would lead to bad results. A search in this genome space would be much more efficient, because the function that translates any solution to an actual room map processes a lot more information and builds a solution that works. As the maps in Figure \ref{Dung} also show, the solutions are also capable of adapting to existing elements of the environment.

\subsection{Exploiting convex representation}

A representation consisting a a list of real parameters is {\em
  convex} if the weighted average of two instances of the
representation is an instance of the representation.  When a
representation is convex, then we may choose example genes and evolve
weight vectors for averaging them.  We call this process {\em convex
  re-representation}.  It restricts the search space to the convex
hull of the example genes.  This both can lower the search dimension
and, if the examples are well chosen, substantially enhance the
richness of the space. An example showing that such enrichment by use
of examples is possible appears \cite{Ashlock18Julia}.

In \cite{Johnson10} the authors use a two-dimensional cellular
automata with a majority based rule to generate cavern-like maps.  A
{\em fashion based cellular automaton} is an automata whose rule is
based on a competition matrix that gives the score that a cell in one
state obtains against a cell in another.  Updating consist of
compiling the scores, based on current neighbors, of each cell and
then adopting the state of the highest scoring member of a cells
neighborhood.  This representation was used to make cavern maps with
diverse appearances.  In \cite{Ashlock19ca, AshlockCA19} pairs of
automata were evolved so that the weighted averages of the pairs all
generate good maps.

\begin{figure*}[tb]
\centerline{\includegraphics[width=0.98\textwidth]{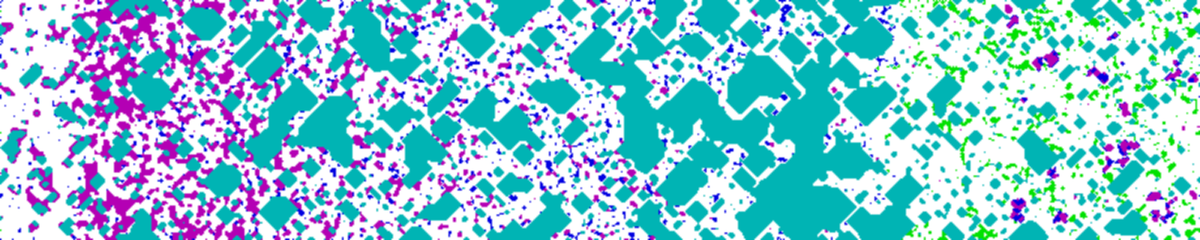}}
\caption{A rendering of a fashion based cellular automata that
  traverses the weighted averages of two rules from left to right.}
\label{Morph}
\end{figure*}

An example demonstrating, via continuously changing weighted average
from left to right, the intermediate rules between two co-evolved
examples if shown in Figure \ref{Morph}.  The representation for
fashion based cellular automata rules is a square matrix or real
parameters, the score values, of size $N\times N$ where $N$ is the
number of states.  The weighted average of two such matrices is such a
matrix.  The morph between two rules demonstrates that convex
re-representation is an excellent way to enrich a search space by
restricting it to a good region while retaining diversity of possible
solutions.

\section{Constructively Rich Spaces}
\label{OP}

\begin{figure}[htb]
\centerline{\includegraphics[width=0.48\textwidth]{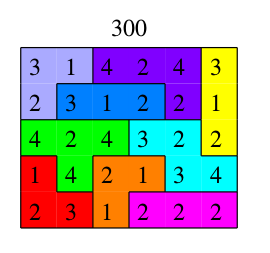}}
\caption{An example of a polyomino math puzzle.}
\label{PMP}
\end{figure}

Figure \ref{PMP} shows an example of a puzzle located in an automatic
content generation effort.  The player is presented with the
polyominos, represented in the figure by colored regions, and a board
with the numbers on it.  To work the puzzle, the player puts the
polyominos on the board.  Their score is the sum of the scores for the
individual polyominos, which are themselves the better of the sum or
product of the numbers under the polyomino.  The score of 300 in the
figure can be verified by the reader and is thought to be optimal.

The software that located these puzzles is as yet unpublished and is a
terrible answer to the problem of finding an enriched search space.
Worse, a majority of the puzzles located achieved their best scores
when some of the pieces were not used -- this is the {\em common}
outcome of procedural content generation on this problem.  This,
however, is not the reason we include these puzzles.

It is clear that, once you know the size of the polyominos, then the
numbers may be optimized by simply assigning them to groups with the
same sizes as the polyominos.  The score can then be optimized by a
search that changes that assignment.  Suppose that we have an optimal
assignment.  Then placement of that assignment under the polyominos
will yield a board where the optimal solution uses all the polyominos.
This still does not give us an enriched space.

First of all, note that the order of the numbers under a given
polyomino is irrelevant to the score.  In addition, once there is one
way to place polyominos to fill a rectangle, there is often a huge
combinatorial space of such placements \cite{Ashlock16gbs}.  This
means that locating one puzzle where the optimal solution uses all the
pieces gives us a combinatorial space of thousands to millions of
equivalent puzzles with very different number layouts. 

This is a constructively rich design space and the idea exemplified here can be
applied to other sorts of puzzles.  A good analogy could be made with
a biological neutral network \cite{Ebner2001OnNN} -- the many forms of the initial puzzle form an enormous neutral network in the search space. This means there is an operator that can take one already found good solution and produce a large range of equally good solutions. This allows one, once a single good solution was found, to define an subspace that only contains good solutions. This means that we have managed once to answer the Riddle of Togelby successfully, albeit in a very limited domain.

\section{Conclusions}

This study has given a number of methods, some quite special purpose
and others general, for creating enriched search spaces for game
content. Based on the example so far, it looks like the rich search spaces we saw so far all shared a relatively complex translation process from their minimal representation to the final artefact. These non-deterministic processes where able to process additional information to ensure the feasibility of the produced artefact, and could often be manipulated in understandable ways to influence the output. This allowed for more human control on the output on and more option to adapt to existing content. 

The other main method discussed here are way to generate subspaces from bigger subspaces, by finding operators that modify representations to produce other artefacts of similar quality, and thereby defining a good subspace, or concentrating of limiting the search to subspaces that have a higher rate of good solutions.

It is to be hoped that this paper gives the games research community some grist for their research mills.  Among the things to do are the following.  Locate more game-content types that are rich as a matter of chance or nature, like the Voronoi path networks.  Work to understand the capabilities and limits of single parent techniques, convex representations, and self-enriching representations like the do-whats-possible representation.  At a lower priority level, construct other spaces like the polyomino math puzzle where one positive example has a billion cousins that are all good games.  Beyond this, there are doubtless other ways to attack the Riddle of Togelby, we hope to hear back that our colleagues have found more of them.

\clearpage

\bibliography{reference}

\end{document}